%% file: root.tex
\documentclass[a4paper, 10pt, conference]{ieeeconf} 
 
\IEEEoverridecommandlockouts%
\usepackage[pdftex]{graphicx}
\graphicspath{{figures/}}
\usepackage{url}
\usepackage{footmisc}
\usepackage{hyperref}
\hypersetup{
	colorlinks=true,
	linkcolor=black,
	citecolor=black,
	filecolor=black,
	urlcolor=black,
}
\usepackage{mathrsfs}
\usepackage{phfparen}
\usepackage{amsmath,amsopn,amssymb}
\usepackage{mathdots}
\usepackage[T1]{fontenc}
\usepackage[utf8]{inputenc}
\usepackage{csquotes}
\usepackage[english]{babel}
\usepackage[export]{adjustbox}
\usepackage{caption, subcaption}
\usepackage{dsfont}
\usepackage[letterpaper,left=.75in,right=.75in,top=.72in,bottom=.8in]{geometry}
\usepackage{mathtools}
\usepackage{calc}
\usepackage{pgfplots}
\usepackage{siunitx}
\usepackage{cases}
\usepackage{tabularx,booktabs}
\newcolumntype{C}{>{\centering\arraybackslash}X} 
\usepackage{placeins}
\usepackage{multirow}
\usepackage{acro}
\usepackage{bm}
\usepackage{upgreek}
\usepackage{makeidx}
\usepackage{newtxtext}
\usepackage{newtxmath}
\usepackage{xcolor,colortbl}
\usepackage{float}
\usepackage[capitalize]{cleveref}
\usepackage{import}

\input{content/acronyms}
\input{content/options/figures}
\input{content/options/definitions}

\usepackage[style=ieee,
            doi=false,
            url=false,
            mincitenames=1,
            maxcitenames=1,
            minbibnames=6,
            maxbibnames=6,
            backend=biber]{biblatex}
\AtBeginBibliography{\small}
\usepackage{todonotes} 

\title{%
\LARGE \bf Fast and Robust Ground Surface Estimation from \acs{LIDAR} Measurements using Uniform B-Splines
}

\author{Sascha Wirges\(^{1,\dagger}\), Kevin Rösch\(^{\ast,2,3}\), Frank Bieder\(^{2,3}\) and
		Christoph Stiller\(^{2,3}\)%
\thanks{\(^1\) Author is with the Bosch Center for Artifical Intelligence, Renningen, Germany.
{\tt\small sascha.wirges@de.bosch.com}}
\thanks{\(^2\) Authors are with the Institute of Measurement and Control Systems, Karlsruhe Institute of Technology (KIT), Karlsruhe, Germany.%
{\tt\small \{kevin.roesch, frank.bieder, stiller\}@kit.edu}}
\thanks{\(^3\) Authors are with the Mobile Perception Systems Group, FZI Research Center for Information Technology, Karlsruhe, Germany.}%
\thanks{\(^\ast\) Corresponding author.}
\thanks{\(^\dagger\) This work was done while Sascha was a PhD candidate at KIT.}
}

\begin{document}
\maketitle

\pubid{\begin{minipage}{\textwidth}~\\[12pt] \centering%
    \copyright~2021 IEEE. Personal use of this material is permitted. Permission from IEEE must be obtained for all other uses, in any current or future media, including reprinting/republishing this material for advertising or promotional purposes, creating new collective works, for resale or redistribution to servers or lists, or reuse of any copyrighted component of this work in other works.
  \end{minipage}}
  \pubidadjcol

  \pagestyle{empty}

\input{content/abstract}

\input{content/introduction}
\input{content/related_work}
\input{content/methodology}
\input{content/experiments}
\input{content/validation}

\input{content/conclusion}
\printbibliography%
\end{document}

%% file: content/acronyms.tex
\usepackage{acro}

\DeclareAcronym{AD}{
    short = AD,
    long = automated driving,
}
\DeclareAcronym{UBS}{
    short = UBS,
    long = uniform B-spline,
}
\DeclareAcronym{LIDAR}{
    short = LiDAR,
    long = light detection and ranging,
}
\DeclareAcronym{LLS}{
    short = LLS,
    long = linear Least-Squares,
}
\DeclareAcronym{LS}{
    short = LS,
    long = Least-Squares,
}
\DeclareAcronym{2D}{
    short = 2D,
    long = two-dimensional,
}
\DeclareAcronym{3D}{
    short = 3D,
    long = three-dimensional,
}
\DeclareAcronym{TLS}{
    short = TLS,
    long = truncated Least-Squares,
}
\DeclareAcronym{OLS}{
    short = OLS,
    long = ordinary Least-Squares,
}\DeclareAcronym{GMC}{
    short = GMC,
    long = Geman McClure,
}
\DeclareAcronym{WLS}{
    short = WLS,
    long = weighted Least-Squares,
}
\DeclareAcronym{GNC}{
    short = GNC,
    long = Graduated Non-Convexity,
}
\DeclareAcronym{GPU}{
    short = GPU,
    long = Graphics Processing Unit,
}

%% file: content/options/figures.tex
\usepackage{tikz}
\usetikzlibrary{arrows, shapes, positioning}

\usepackage[inkscapepath=out/svg-inkscape/, notransparent]{svg}
\svgpath{components/figures/}

\graphicspath{{components/figures/}}

\usepackage{pgfplots}

\usepgfplotslibrary{
    colorbrewer,
    groupplots,
}

\pgfplotsset{
    bar cycle list/.style={
            cycle list name=Set1,
            every axis plot/.append style={fill, thin},
        },
    compat=1.15,
    colormap/Spectral,
    colormap={reverse Spectral}{
            indices of colormap={
                    \pgfplotscolormaplastindexof{Spectral},...,0 of Spectral}
        },
    cycle list/Set1,
    cycle list name=Set1,
    enlarge x limits=0.05,
    enlarge y limits=0.05,
    every axis/.append style={semithick},
    every axis plot/.append style={thick},
    grid style={
            opacity=0.5,
        },
    height=0.25\linewidth,
    scale only axis=true,
    scaled x ticks=false,
    scaled y ticks=false,
    title style={
            anchor=north east,
            at={(0.99, 0.95)},
            fill=white,
            fill opacity=0.5,
            text opacity=1,
        },
    width=0.9\linewidth,
}

\newrobustcmd*{\tribetter}{\tikz{\filldraw[draw=black!30!green,fill=black!30!green] (0,0) -- (1mm,0) -- (0.5mm,1mm);}}
\newrobustcmd*{\triworse}{\tikz{\filldraw[draw=black!30!red,fill=black!30!red] (0,0) -- (1mm,0) -- (0.5mm,1mm);}}

%% file: content/options/definitions.tex
\newcommand{\func}[1]{\mathrm{#1}}
\newcommand{\mat}[1]{\boldsymbol{#1}}
\newcommand{\matfunc}[1]{\bm{\mathrm{#1}}}

\DeclarePairedDelimiterX{\inp}[2]{\langle}{\rangle}{#1, #2}
\DeclarePairedDelimiter\norm{\lVert}{\rVert}
\DeclarePairedDelimiter\abs{\lvert}{\rvert}

\DeclareMathOperator*{\argmin}{arg\,min}
\DeclareMathOperator{\diag}{diag}

%% file: content/abstract.tex
\begin{abstract}
    We propose a fast and robust method to estimate the ground surface from \acs{LIDAR} measurements on an automated vehicle.
    The ground surface is modeled as a \acl*{UBS} which is robust towards varying measurement densities and with a single parameter controlling the smoothness prior.
    We model the estimation process as a robust \acl*{LS} optimization problem which can be reformulated as a linear problem and thus solved efficiently.
    Using the SemanticKITTI data set, we conduct a quantitative evaluation by classifying the point-wise semantic annotations into ground and non-ground points.
    Finally, we validate the approach on our research vehicle in real-world scenarios.

\end{abstract}

%% file: content/introduction.tex

\section{Introduction}\label{sec:introduction}
Moving robotic platforms such as automated vehicles need to make safe, reasonable and quick decisions based on their surroundings.
Therefore, the environment perception and modeling has to provide an accurate representation of the surroundings at a high frequency.
Clearly the level of abstraction and the geometric representation of the environmental modeling is highly dependent on the task, domain and safety characteristics of the application.
However, one key component of meaningful environment representations is an accurate ground surface estimation.
It is required to extract necessary features towards a holistic scene understanding such as the height of objects and information about the driving areas.

\begin{figure}
    \begin{tikzpicture}
        \begin{groupplot}[
                axis on top,
                major grid style={very thin},
                enlargelimits=0,
                grid=major,
                group style={
                        group size=1 by 3,
                        horizontal sep=1mm,
                        vertical sep=2mm,
                        xlabels at=edge bottom,
                        ylabels at=edge left,
                    },
                height=0.34\linewidth,
                title style={
                        anchor=north east,
                        at={(0.98,0.95)},
                        fill=white,
                    },
                width=0.85\linewidth,
                xlabel={x / \si{\metre}},
                xtick={-45, -30, -15, ..., 45.0001},
                xticklabels={-45, -30, -15, ..., 45},
                ylabel={y / \si{\metre}},
                ytick={-30, -15, ..., 30.0001},
                yticklabels={-30, -15, ..., 30},
            ]
            \nextgroupplot[
                title=Ground surface height,
                xticklabels={,,},
            ]
            \addplot graphics[xmin=-50,xmax=50,ymin=-20,ymax=20] {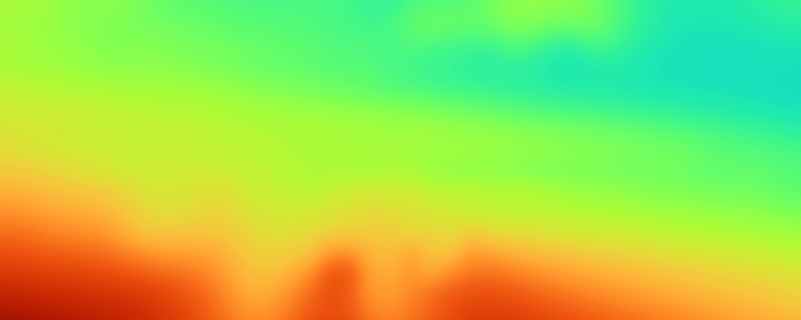};
            \addplot graphics[xmin=-2.2,xmax=3,ymin=-1.9,ymax=1.9] {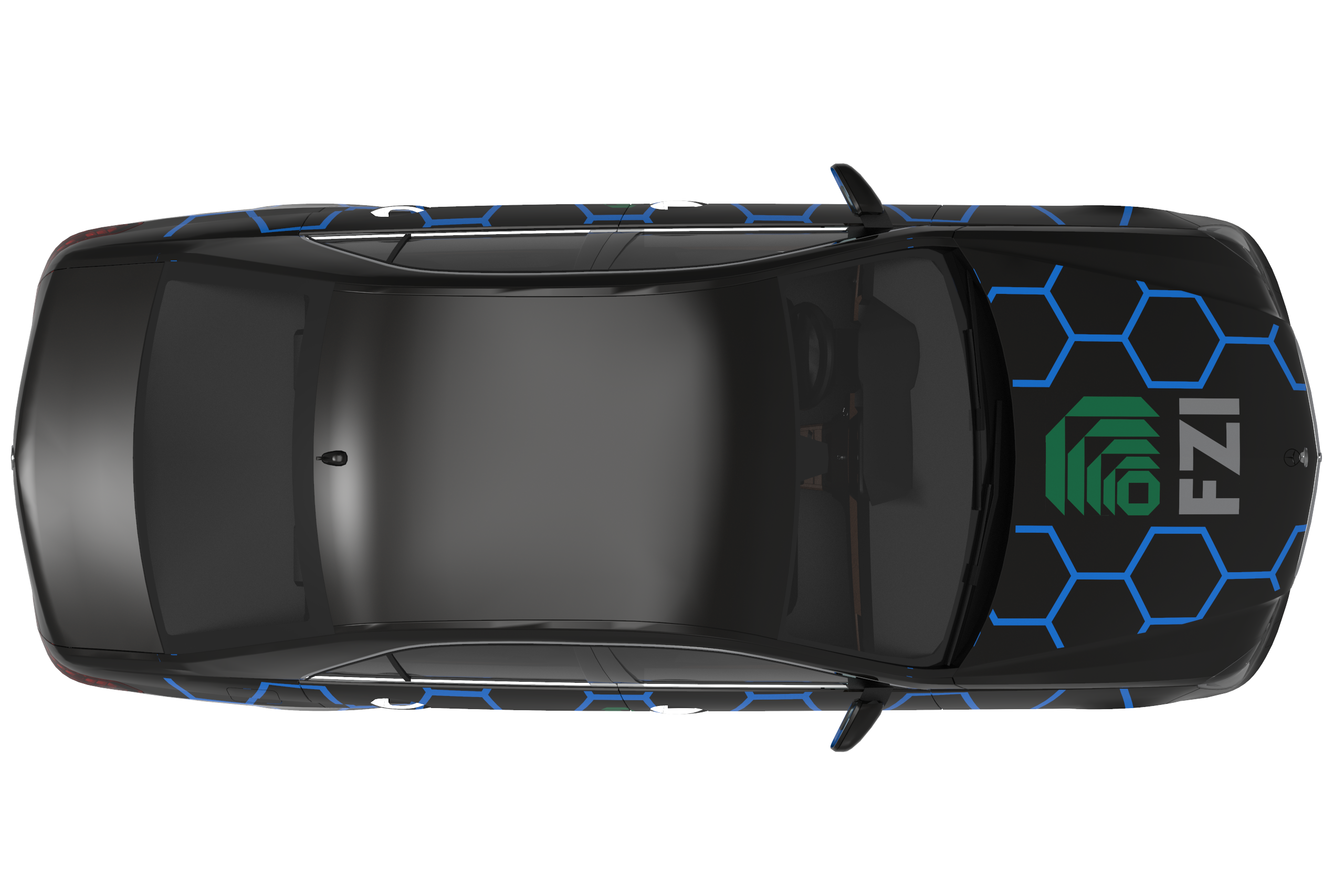};
            \nextgroupplot[
                title=Height limit,
                xticklabels={,,},
            ]
            \addplot graphics[xmin=-50,xmax=50,ymin=-20,ymax=20] {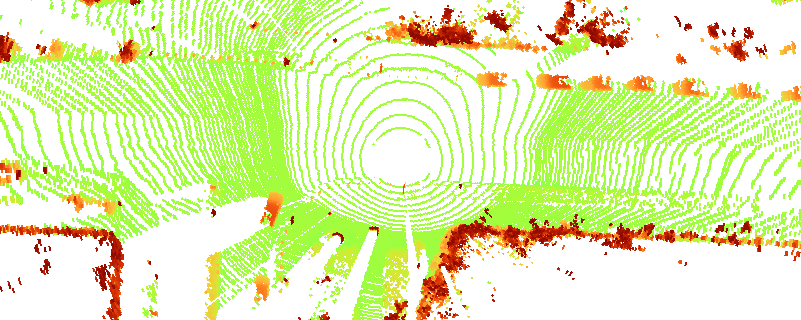};
            \addplot graphics[xmin=-2.2,xmax=3,ymin=-1.9,ymax=1.9] {fig/diss/top};
            \nextgroupplot[
                title=Combined height map,
            ]
            \addplot graphics[xmin=-50,xmax=50,ymin=-20,ymax=20] {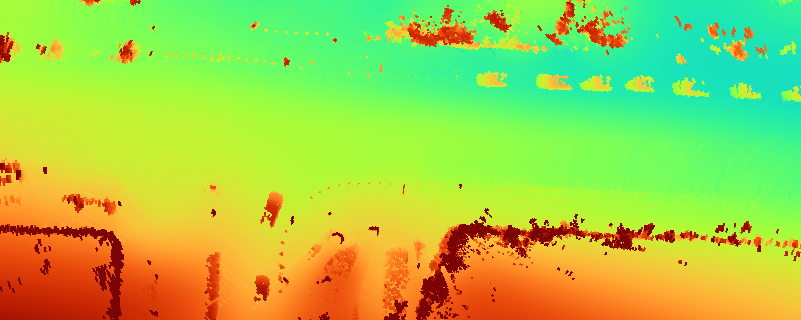};
            \addplot graphics[xmin=-2.2,xmax=3,ymin=-1.9,ymax=1.9] {fig/diss/top};
        \end{groupplot}
    \end{tikzpicture}
    \caption{
        The estimated ground surface (top) is the result of the presented approach.
        In a subsequent processing step, we can then use the max.~observed reflection height in each grid cell (center) to construct a combined height map (bottom).
        White/blue/red: No/low/high values.
    }\label{fig:gm_height_mapping}
\end{figure}
To accomplish scene understanding, we require geometric information on the ground surface.
The ground surface can be obtained from a map, estimated from range sensor measurements or combined from multiple sources.
Using map information has the disadvantage that an accurate pose estimate (incl.~roll and pitch) needs to be available which can not be guaranteed at all times.
In consequence, we follow the approach of estimating the ground surface from range sensor measurements to be independent to other sources of errors (e.g., ~from pose estimation).
To achieve the fast and robust ground surface representation we utilize \acl*{UBS}.
With the ground surface model \(\func{g}\colon \mathbb{R}^2 \rightarrow \mathbb{R}\) we map the plane coordinates from the measurement points to the plane distance.

\renewcommand{\thefootnote}{\fnsymbol{footnote}}
\pubidadjcol
We provide a quantitative evaluation based on the SemanticKITTI data set, classifying \acs{LIDAR} measurements into ground and non-ground.
In addition, we validate our approach on our research vehicle BerthaOne \cite{Tacs2018Making}, based on range measurements of five \acsp{LIDAR}.
The source code is available on github.\footnote[7]{\url{https://github.com/KIT-MRT/pointcloud\_surface}}


%% file: content/related_work.tex
\section{Related Work}\label{sec:related_work}

Instead of estimating the ground surface shape, \textcite{Moosmann2009} segment each point in range images using a local convexity criterion.
The method provides accurate point classification results, but it is only possible to apply this method to single range measurements in an image structure.
Thus, it is hard to resolve point classification conflicts when multiple measurements are available.

\Textcite{Zhang2003} develop a progressive morphological filter to estimate the ground surface in airborne LiDAR measurements represented on an elevation grid.
By gradually increasing the filter window size and using elevation thresholds, the authors remove non-ground measurements while preserving the ground surface elevation.
Their method works well on accurate elevation maps with large ground areas and local elevations such as buildings or trees.
However, \textcite{Zhang2003} do not consider measurement errors such as multi-path propagation which would lead to false elevation estimates.
In addition, the approach is not real-time capable as the filtering has to be applied several times.

A popular representation for ground surfaces are polynomials
\begin{equation}
    \begin{split}
        \func{h}`*(\mat{x}) & = \func{h}`*(x_1, x_2) \\
        & = w_0 + w_1 x_1 + w_2 x_2 + w_3 x_1 x_2 + w_4 x_1^2 + \ldots\\
        & = \inp*{\mat{w}}{\mat{x}}
    \end{split}
\end{equation}
such as planes, quadratics or cubics which can be expressed as a linear combination of weights \(\mat{w}\) and a transformed input \(\mat{x}\).
This yields a system of linear equations which can be solved efficiently, e.g.~by using a \ac{LLS} method.
For example, \textcite{Saleem2017} use a polynomial representation to fit a ground surface on v-disparity estimates of stereo cameras.

\Textcite{Wedel2009} model the ground surface along the driving direction by a univariate B-spline. 

They estimate and track its parameters in a \ac{LLS} approach combined with a Kalman filter.

\Textcite{Beck2020} uses \ac{UBS} surfaces with smoothness regularization to model viewing rays depending on camera image coordinates.
He describes smoothness of the \(n\)-th derivative by the penalty
\begin{equation}
    \rho = \int_{0}^1 \norm*{\func{s}^{`*(n)}`*(x)}^2 \mathrm{d}x = \norm*{\matfunc{f}_s`*(\mat{p})}_F^2\label{eq:smoothness_term}
\end{equation}
and shows that its closed-form solution
\begin{equation}
    \matfunc{f}_s`*(\mat{p}) = \mat{B}_s~\mat{p}, \quad \mat{B}_s = \mat{B}_d `*(\int_0^1 \matfunc{a}^{`*(n)}`*(x)~\matfunc{a}^{\top `*(n)}`*(x) \mathrm{d}x)^{\frac{1}{2}}\label{eq:smoothness_term_1d}
\end{equation}
is a matrix-vector product of the control points \(\mat{p}\) and the term \(\mat{B}_s\) which only depends on the spline degree and the smoothness derivative \(n\) so that it can be precomputed.

%% file: content/methodology.tex
\section{Methodology}

\subsection{Problem Formulation}\label{sec:problem_formulation}

In this work, we use \acp{UBS} to model the ground surface.
Due to their local support, splines are robust towards varying measurement densities as it is often the case for range sensors.
However, it is still possible to impose smoothness constraints on splines to reduce overfitting, especially in areas with few measurements.

Given \(N\) positions \(\mat{x}_1, \ldots, \mat{x}_N\), we can denote ground surface height estimates in the form
\begin{align}
    \hat{h}_n     & = \hat{\func{h}}`*(\mat{x}_n, \mat{p}) = \inp{\matfunc{b}`*(\mat{x}_n)}{\mat{p}}~, \quad \matfunc{b}\colon \mathbb{R}^{2} \rightarrow \mathbb{R}^{N_C} \\
    \hat{\mat{h}} & = \hat{\matfunc{h}}`*(\mat{p}) =
    \begin{bmatrix}
        \matfunc{b}^\top`*(\mat{x}_1) \\
        \matfunc{b}^\top`*(\mat{x}_2) \\
        \vdots                        \\
        \matfunc{b}^\top`*(\mat{x}_N)
    \end{bmatrix}
    \mat{p} = \mat{B}~\mat{p}
\end{align}
with the spline control points \(\mat{p} \in \mathbb{R}^{N_C}\) using a vectorial basis function \(\matfunc{b}\) weighting \(N_C\) control points depending on the \acs{2D} positions \(\mat{x}_n\).

Then, given \(N\) pairs \(`*(\mat{x}_n, h_n)\) of positions and height measurements we aim to find parameters
\begin{equation}
    \mat{p}^* = \argmin_{\mat{p}} \sum_{n=1}^N \uprho`*(`*(\hat{\func{h}}`*(\mat{x}_n, \mat{p}) - h_n)^2)
\end{equation}
that minimize the sum of squared and robustified errors.
Here, we use the non-linear robustifier \(\uprho\colon \mathbb{R} \rightarrow \mathbb{R}^+\) as the optimization needs to be robust against outliers, i.e.~non-ground points contained in the measurements.
Assuming that the conditions on the robustifier \(\uprho\) hold, we can formulate the equivalent dual \ac{WLS} problem
\begin{equation}
    \mat{p}^* = \argmin_{\mat{p}, w_1, \ldots, w_N} \sum_{n=1}^N w_n `*(\hat{\func{h}}`*(\mat{x}_n, \mat{p}) - h_n)^2 + \upPhi`*(w_n)
\end{equation}
according to the Black-Rangarajan duality \cite{Black1996}.
The weight penalty \(\upPhi\colon \mathbb{R}^+ \rightarrow \mathbb{R}^+\) prevents the weights to become zero.

We regularize the spline towards constant slope by penalizing variations of the second spline derivative.
In absence of measurements, this will lead to an extrapolation with constant incline.
As presented by \textcite{Beck2020}, the closed-form solution of the smoothness cost term is a linear combination of the control points with precomputed weights \(\mat{B}_\text{S}\).
This yields the final optimization problem
\begin{align}
    \mat{p}^*, w_1^*, \ldots, w_N^* = \argmin_{\mat{p}, w_1, \ldots, w_N} & \sum_{n=1}^N w_n `*(\inp{\matfunc{b}`*(\mat{x}_n)}{\mat{p}} - h_n)^2 \nonumber \\+ \upPhi`*(w_n)
                                                                          & + w_{\text{S}} \norm*{\mat{B}_\text{S} \mat{p}}^2,
\end{align}
including a weighted spline cost with weight penalty \(\upPhi\) and the smoothness cost weighted by \(w_\text{S}\).
By reordering indices, this problem can be formulated by the \ac{WLS} problem
\begin{equation}
    \mat{p}^*, \mat{w}^* = \argmin_{\mat{p}, \mat{w}} \norm*{\diag`*(\tilde{\mat{w}})^{\frac{1}{2}} `*(\tilde{\mat{B}} \mat{p} - \tilde{\mat{h}})}^2 + \sum_{n=1}^N \upPhi`*(w_n)\label{eq:gse_problem}
\end{equation}
with
\begin{align}
    \tilde{\mat{B}} = \begin{bmatrix} \mat{B}\\ \mat{B}_{\text{S}} \end{bmatrix}, \quad \tilde{\mat{h}} = \begin{bmatrix} \mat{h}\\ \mat{0} \end{bmatrix}, \nonumber \\ \tilde{\mat{w}} = \begin{bmatrix}w_{1} & \ldots & w_{N} & w_S & \ldots & w_S\end{bmatrix}^\top.
\end{align}

\subsection{Parameter Estimation}\label{sec:parameter_estimation}
To solve \cref{eq:gse_problem}, we use the iterative \ac{GNC} method presented by \textcite{Yang2020}.
The approach uses weight penalty functions \(\upPhi_\mu\) with a free parameter \(\mu\) which controls problem convexity.
The algorithm repeats two steps in an alternating fashion while changing \(\mu\) in order to decrease convexity.
In step 1, we fix the weights \(\mat{w}\) and determine the optimal control points
\begin{align}
    \mat{p}^* & = \argmin_{\mat{p}} \norm*{\diag`*(\mat{w})^{\frac{1}{2}} `*(\tilde{\mat{B}} \mat{p} - \tilde{\mat{h}})}^2            \nonumber \\
              & = `*(\tilde{\mat{B}}^\top \diag`*(\mat{w}) \tilde{\mat{B}})^{-1} \tilde{\mat{B}}^\top \diag`*(\mat{w}) \tilde{\mat{h}}
\end{align}
of the resulting linear \ac{WLS} problem.
In step 2, we fix the parameters \(\mat{p}\) and determine the weights
\begin{align}
    \mat{w}^* & = \argmin_{\mat{w}} \sum_{n=1}^N w_n `(\overbrace{\inp{\matfunc{b}`*(\mat{x}_n)}{\mat{p}} - h_n}^{\Delta h_n})^2 +  \upPhi`*(w_n) \\
    w_n^*     & = \argmin_{w_n} w_n \Delta h_n^2 + \upPhi`*(w_n)
\end{align}
independently of each other.
Here, we investigate the \ac{GMC} and \ac{TLS} penalties
\begin{align}
    \upPhi_{\mu, \text{GMC}}`*(w) & = \mu c^2 `*(\sqrt{w}-1)^2        \\
    \upPhi_{\mu, \text{TLS}}`*(w) & = \frac{\mu`*(1 - w)}{\mu + w}c^2
\end{align}
for which we can compute the optimal weights
\begin{align}
    w_{n, \text{GMC}}^* & = `*(\frac{\mu c^2}{\mu c^2 + \Delta h_n^2})^2\label{eq:opt_w_gmc} \\
    w_{n, \text{TLS}}^* & =
    \begin{dcases*}
        1                                                       & if \(\Delta h_n^2 < \frac{\mu}{\mu + 1}c^2\)                                \\
        \frac{c \sqrt{\mu`*(\mu + 1)}}{\abs*{\Delta h_n}} - \mu & if \(\frac{\mu}{\mu + 1}c^2 \leq \Delta h_n^2 \leq \frac{\mu + 1}{\mu}c^2\) \\
        0                                                       & otherwise
    \end{dcases*}\label{eq:opt_w_tls}
\end{align}
in closed form.

We adapt the convexity parameter \(\mu^{`*(k)}\) in every iteration \(k\) such that the next
\begin{equation}
    \mu^{`*(k+1)} = \alpha \mu^{`*(k)}
\end{equation}
is changed by a constant factor \(\alpha\).
This factor is set to \(\alpha=1.6^{-1}\) for the \ac{GMC} penalty and \(\alpha=1.6\) for the \ac{TLS} penalty.
The optimization is stopped after a fixed number of steps or if \(\mu^{`*(k+1)} < 1\).

As the measurements contain outliers (non-ground points) with a biased distribution, the method overestimates the true ground surface in general.
To mitigate this issue, we scale positive and negative errors \(\Delta h_n\) differently.
If \(\Delta h_n > 0\), i.e.~the point lies above the current ground estimate, we scale it with an asymmetry ratio \(r_{\text{asymm}} > 1\) yielding the asymmetric error
\begin{equation}
    \Delta \tilde{h}_n =
    \begin{dcases*}
        r_{\text{asymm}}~\Delta h_n & if \(\Delta h_n > 0\)    \\
        \Delta h_n                  & otherwise
    \end{dcases*},
\end{equation}
which we use to substitute \(\Delta h_n\) in \cref{eq:opt_w_gmc,eq:opt_w_tls}.
Due to the error scaling, points above the current ground estimate may get a lower weight because of the higher distance.
In other words, points below the current estimate are more likely to belong to the ground.

\begin{figure}[!t]
    \begin{tikzpicture}[
            declare function={
                    asym(\x,\r) = (\x >= 0) * \r * \x + (\x < 0) * \x;
                    gmc(\x,\c,\mu,\r) = ((\mu * \c^2) / (\mu * \c^2 + asym(\x,\r)^2))^2;
                    tls_lb(\c,\mu) = (\mu * \c^2) / (\mu + 1);
                    tls_ub(\c,\mu) = (\mu + 1) * \c^2 / \mu;
                    tls(\x,\c,\mu,\r) = (asym(\x,\r)^2 < tls_lb(\c,\mu)) * 1 + and(asym(\x,\r)^2 >= tls_lb(\c,\mu), asym(\x,\r)^2 < tls_ub(\c,\mu)) * (\c / abs(asym(\x,\r)) * sqrt(\mu * (\mu + 1)) - \mu) + (asym(\x,\r)^2 >= tls_ub(\c,\mu)) * 0;
                },
        ]
        \begin{groupplot}[
                domain=-0.85:0.55,
                enlarge x limits=0,
                group style={
                        group size=1 by 2,
                        vertical sep=0.25cm,
                        xlabels at=edge bottom,
                    },
                legend style={
                        at={(current bounding box.south)},
                        anchor=north,
                        legend columns=-1
                    },
                xlabel={\(\Delta h\) / \si{\metre}},
                ymin=0,
                ytick={0, 0.25, ..., 1},
            ]
            \nextgroupplot[
                width=0.8\linewidth,
                samples=200,
                xticklabels={,,},
                ylabel={\(w^{`*(k)}\) (GMC)},
            ]
            \draw[style=dotted] (-1, 0) -- (1, 0);
            \draw[style=dotted] (0, -0.1) -- (0, 1.1);
            \draw[style=dotted] (0.2, -0.1) -- (0.2, 1.1);
            \draw[style=dotted] (-0.4, -0.1) -- (-0.4, 1.1);
            \addplot+ {gmc(x,0.4,68.75,2)};
            \addplot+ {gmc(x,0.4,26.86,2)};
            \addplot+ {gmc(x,0.4,10.49,2)};
            \addplot+ {gmc(x,0.4,4.10,2)};
            \addplot+ {gmc(x,0.4,1.6,2)};

            \nextgroupplot[
                width=0.8\linewidth,
                samples=300,
                ylabel={\(w^{`*(k)}\) (TLS)},
            ]
            \draw[style=dotted] (-1, 0) -- (1, 0);
            \draw[style=dotted] (0, -0.1) -- (0, 1.1);
            \draw[style=dotted] (0.2, -0.1) -- (0.2, 1.1);
            \draw[style=dotted] (-0.4, -0.1) -- (-0.4, 1.1);
            \addplot+ {tls(x,0.4,1,2)};
            \addplot+ {tls(x,0.4,2.56,2)};
            \addplot+ {tls(x,0.4,6.56,2)};
            \addplot+ {tls(x,0.4,16.79,2)};
            \addplot+ {tls(x,0.4,42.97,2)};
            \legend{\(k=0\), \(k=2\), \(k=4\), \(k=6\), \(k=8\)}
        \end{groupplot}
    \end{tikzpicture}
    \caption{
        Weights of \ac{GMC} and \ac{TLS} penalty functions for \(c=\SI{0.4}{\metre}\) and \(r_\text{asymm}=2\) at different convexity parameters \(\mu^{`*(k)}\) changing during optimization.
    }\label{fig:penalty_weights_asymmetric}
\end{figure}
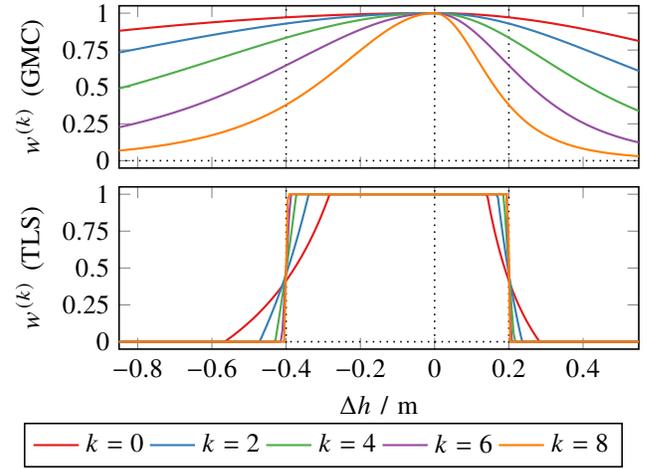

\Cref{fig:penalty_weights_asymmetric} depicts the optimal weights as a function of the height difference \(\Delta h\) for different \(\mu^{`*(k)}\), i.e.~at different optimization steps.
Due to the change of the convexity parameter \(\mu\), measurements with larger errors yield smaller weights, thus gain less influence on the total cost.

%% file: content/experiments.tex
\section{Experiments}\label{sec:experiments}

We evaluate on the training set of the SemanticKITTI data set \cite{Behley2019} and split all range measurements into ground, non-ground and don't care classes, summarized in \cref{tab:kitti_ground_classes}.

\begin{table*}
    \begin{center}
        \begin{tabular}{c m{80mm} r r}
            \toprule
            \textbf{Category} & \textbf{Classes}                                                                                                                     & \textbf{Abs.~frequency} & \textbf{Rel.~frequency} \\
                              &                                                                                                                                      & in million              & in \%                   \\
            \midrule
            Ground            & Lane marking, Other, Parking, Road, Sidewalk, Terrain                                                                                & \num{10196.462}         & \num{43.949}            \\
            \midrule
            Non-ground        & Bicycle, Bicyclist, Building, Bus, Car, Fence, Motorcycle, Motorcyclist, Other, Person, Pole, Traffic sign, Truck, Trunk, Vegetation & \num{12559.459}         & \num{54.133}            \\
            \midrule
            Don't care        & Unlabeled, Outlier                                                                                                                   & \num{445.079}           & \num{1.918}             \\
            \midrule
            Total             &                                                                                                                                      & \num{23201.000}         & \num{100.000}           \\
            \bottomrule
        \end{tabular}
    \end{center}
    \caption{
        Distribution of ground, non-ground and don't care classes constructed from the SemanticKITTI training data set.
    }\label{tab:kitti_ground_classes}
\end{table*}

The parameters and default values of our method are summarized in \cref{tab:parameters}.

\begin{table}
    \begin{center}
        \begin{tabular}{l c l}
            \toprule
            \textbf{Parameter}                    & \textbf{Default Value} & \textbf{Related Exp.}            \\
            \midrule
            Ground surface model                  & \Ac{UBS}               & \Cref{subsec:models}         \\
            \midrule
            Robustifier                           & \acs{TLS}              & \Cref{subsec:robustifier}    \\
            Error threshold \(c\)                   & \SI{0.4}{\metre}       & \Cref{subsec:robustifier}    \\
            Initial convexity \(\mu_0\)             & \num{1}                &                                  \\
            Number of iterations                  & \num{10}               &                                  \\
            Asymmetry ratio \(r_{\text{asymm}}\)    & \num{2}                & \Cref{subsec:asymmetry}      \\
            \midrule
            Spline degree                         & \num{2}                &                                  \\
            Control point distance \(d_{\text{C}}\) & \SI{2}{\metre}         & \Cref{subsec:control_points} \\
            Smoothness weight \(w_{\text{S}}\)      & \num{1}                & \Cref{subsec:control_points} \\
            Smoothness order                      & \num{2}                &                                  \\
            \bottomrule
        \end{tabular}
    \end{center}
    \caption{
        Parameters of our ground surface estimation method and their default values.
    }\label{tab:parameters}
\end{table}

\subsection{Comparison of Different Ground Surface Models}\label{subsec:models}
\begin{figure}
    \begin{tikzpicture}
        \begin{groupplot}[
                enlarge x limits=0,
                group style={
                        group size=2 by 1,
                        horizontal sep=0.9cm,
                        xlabels at=edge bottom,
                        ylabels at=edge left,
                    },
                legend style={
                        at={(-0.4, -0.6)},
                        anchor=north west,legend columns=-1, font=\scriptsize},
                ybar=1pt,
                ylabel={Abs.~error / \si{\cm}},
                ymin=0,
            ]
            \nextgroupplot[
                bar width=6pt,
                width=0.12\linewidth,
                symbolic x coords={Average},
                xlabel={Average},
                xmajorticks=false,
                ymax=0.25,
                ytick={0, 0.05, ..., 0.2501},
                yticklabels={0, 5, ..., 25},
            ]
            \addplot+ coordinates {(Average,0.0214)};
            \addplot+ coordinates {(Average,0.0851)};
            \addplot+ coordinates {(Average,0.1301)};
            \addplot+ coordinates {(Average,0.2088)};
            \nextgroupplot[
                bar width=3pt,
                width=0.6\linewidth,
                xlabel={Distance / \si{\metre}},
                xmajorgrids,
                xmax=52,
                xmin=-2,
                xtick={0, 5, ..., 50},
                ymax=1.0,
                ytick={0, 0.2, ..., 1.01},
                yticklabels={0, 20, ..., 100},
            ]
            \addplot+[bar shift=1] coordinates {(0,0.01863) (5,0.01864) (10,0.02215) (15,0.02521) (20,0.02945) (25,0.0342) (30,0.03805) (35,0.04333) (40,0.04947) (45,0.05766)};
            \addplot+[bar shift=2] coordinates {(0,0.06218) (5,0.06662) (10,0.09788) (15,0.11668) (20,0.1363) (25,0.15022) (30,0.15984) (35,0.20372) (40,0.2945) (45,0.37674)};
            \addplot+[bar shift=3] coordinates {(0,0.09407) (5,0.10072) (10,0.14832) (15,0.17394) (20,0.20679) (25,0.24182) (30,0.28272) (35,0.36184) (40,0.48994) (45,0.55175)};
            \addplot+[bar shift=4] coordinates {(0,0.10829) (5,0.15218) (10,0.26334) (15,0.30403) (20,0.3782) (25,0.44264) (30,0.52125) (35,0.63861) (40,0.79191) (45,0.84607)};
            \legend{\Ac{UBS}, Polynomial, Estimated plane, Calibrated plane}
        \end{groupplot}
    \end{tikzpicture}
    \caption{
        Abs.~ground point error of different surface models when only ground points are used for optimization.
        Left: Average of all validation points.
        Right: Averaged within \SI{5}{\metre} intervals of distance from the sensor.
    }\label{fig:abs_error_methods}
\end{figure}
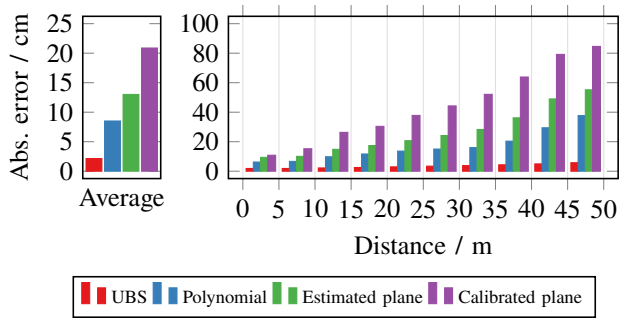

We first compare the accuracy of different ground surface models when only ground points are used for estimation (outlier-free case).
Here, we compare our \ac{UBS} model to cubic polynomials, estimated and precalibrated ground planes.
We randomly sample 10 \% of all ground points for validation, i.e.~these points are not used during optimization.
We then compare the absolute height error between all validation points and the ground surface height estimated by the models.
\Cref{fig:abs_error_methods} depicts the average absolute height errors and average errors depending on measurement distance.

In general, the \ac{UBS} model maintains the lowest errors.
Compared to the polynomial model the error only slightly increases with increasing measurement distances as the influence of measurements to the \ac{UBS} model is restricted locally and thus is almost independent of locally varying measurement densities.

\subsection{Comparison of Robustifiers}\label{subsec:robustifier}

\begin{figure}
    \begin{tikzpicture}
        \begin{axis}[
                ybar=0pt,
                bar width=0.07,
                bar shift=0.05,
                height=0.25\linewidth,
                width=0.85\linewidth,
                xlabel={Ground distance / m},
                xmajorgrids,
                xtick={-0.5, 0, ..., 4.0},
                ylabel={Relative frequency / \%},
                ymax=0.055,
                ymin=0,
                ytick={0, 0.01, ..., 0.04, 0.05},
                yticklabels={0, 1, ..., 5},
            ]
            \addplot+ coordinates {(-0.5,0.000975087666810314) (-0.4,0.0013267177081459365) (-0.3,0.001962106328431433) (-0.2,0.0035280679698819492) (-0.1,0.014804477590917301) (0.0,0.05213341045243105) (0.1,0.04820588445578884) (0.2,0.045829209488240256) (0.3,0.0479400000820163) (0.4,0.04978137009185172) (0.5,0.05078393024858131) (0.6,0.05044772283814683) (0.7,0.050756428042610206) (0.8,0.048522122679391794) (0.9,0.045002598074411725) (1.0,0.04148999730950867) (1.1,0.037875351622612324) (1.2,0.03647897948897068) (1.3,0.03694078179247581) (1.4,0.03617138863645103) (1.5,0.034027865811581334) (1.6,0.03262699169608704) (1.7,0.031754023241399594) (1.8,0.030154096152704372) (1.9,0.027230734979590717) (2.0,0.023133589759087806) (2.1,0.019080899752126526) (2.2,0.01565076103105904) (2.3,0.013469454246284141) (2.4,0.01181473075800953) (2.5,0.010310281343852578) (2.6,0.009064851895494826) (2.7,0.007979458244281755) (2.8,0.007021279010964213) (2.9,0.005816278251124767) (3.0,0.0045615476785434) (3.1,0.003537205655820486) (3.2,0.0028289532764413576) (3.3,0.002210413657544968) (3.4,0.001795369561598493) (3.5,0.0014184957342310435) (3.6,0.0011549143544755316) (3.7,0.0009503044795818928) (3.8,0.0007881885588089284) (3.9,0.0006636783016301691)};
        \end{axis}
    \end{tikzpicture}
    \caption{
        Ground distance histogram of non-ground points.
    }\label{fig:outlier_dist}
\end{figure}
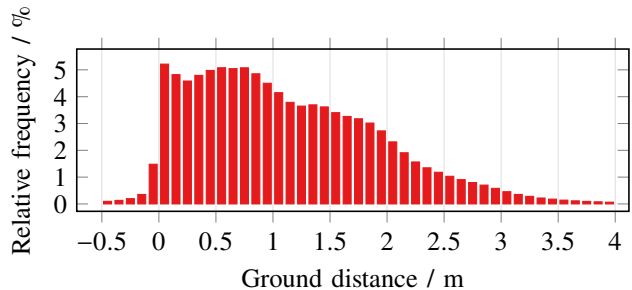

\paragraph{Outlier Noise Estimation}
To justify the use of robust optimization methods, we first aim to estimate the ground distance distribution of non-ground points, i.e.~the outlier noise.
To estimate the outlier noise, we compute ground surfaces only based on the labeled ground points and compute a histogram of the errors between the estimated ground height and the non-ground points, which is depicted in \cref{fig:outlier_dist}.

We observe from the histogram that the non-ground points are highly biased with a mean of \SI{1.09}{\metre} and with a large height range from below the estimated surface to around \SI{4}{m}.
Note that the negative ground distances are likely due to the control point distance of \SI{2}{m} and the smoothness weight of \num{1} which on the one hand reduces overfitting on the inliers but on the other hand increases smoothing of abrupt changes of the ground surface.

\paragraph{Results}

\Cref{fig:robustifier_influence} compares the influence of different robustifiers on the optimization in two settings.
On the one hand, we show the influence when only ground points are used, on the other hand when all points are used for optimization.
For validation, we retain 10\% of ground points in both experiments.

\begin{figure*}
    \begin{tikzpicture}
        \begin{groupplot}[
                enlarge x limits=0,
                group style={
                        group size=2 by 2,
                        horizontal sep=0.9cm,
                        vertical sep=0.3cm,
                        xlabels at=edge bottom,
                        ylabels at=edge left,
                    },
                legend style={
                        anchor=north west,
                        at={(0.03,0.93)},
                        legend columns=3,
                        nodes={scale=0.75, transform shape},
                    },
                ybar=0.5pt,
                ylabel={Abs.~error / \si{\cm}},
                ymin=0,
            ]
            \nextgroupplot[
                bar width=4.5pt,
                width=0.12\linewidth,
                height=0.15\textheight,
                symbolic x coords={Average,},
                xmajorticks=false,
                ymax=0.12,
                ytick={0, 0.02, ..., 0.12},
                yticklabels={0, 2, ..., 12},
            ]
            \addplot+ coordinates {(Average,0.0214)}; 
            \addplot+ coordinates {(Average,0.0212)}; 
            \addplot+ coordinates {(Average,0.0246)}; 
            \addplot+ coordinates {(Average,0.0215)}; 
            \addplot+ coordinates {(Average,0.0372)}; 
            \addplot+ coordinates {(Average,0.0254)}; 
            \addplot+ coordinates {(Average,0.0724)}; 
            \addplot+ coordinates {(Average,0.0499)}; 
            \addplot+ coordinates {(Average,0.0791)}; 

            \nextgroupplot[
                width=0.74\linewidth,
                height=0.15\textheight,
                bar width=0.4,
                xmax=51,
                xmajorgrids,
                xmin=-1,
                xtick={0, 5, ..., 50.0001},
                xticklabels={, ,},
                ymax=0.48,
                ytick={0, 0.05, ..., 0.4501},
                yticklabels={0, , 10, , 20, , 30, , 40, },
            ]
            \addplot+[bar shift=0.5] coordinates {(0,0.01863) (5,0.01864) (10,0.02215) (15,0.02521) (20,0.02945) (25,0.0342) (30,0.03805) (35,0.04333) (40,0.04947) (45,0.05766)}; 
            \addplot+[bar shift=1] coordinates {(0,0.0185) (5,0.01846) (10,0.02172) (15,0.02471) (20,0.02842) (25,0.03287) (30,0.03656) (35,0.04312) (40,0.05649) (45,0.07273)}; 
            \addplot+[bar shift=1.5] coordinates {(0,0.02034) (5,0.0205) (10,0.02553) (15,0.03066) (20,0.03643) (25,0.04445) (30,0.04861) (35,0.0573) (40,0.0703) (45,0.07825)}; 
            \addplot+[bar shift=2] coordinates {(0,0.01849) (5,0.01848) (10,0.02205) (15,0.02519) (20,0.0289) (25,0.03349) (30,0.03956) (35,0.04827) (40,0.06341) (45,0.07885)}; 
            \addplot+[bar shift=2.5] coordinates {(0,0.02488) (5,0.02567) (10,0.0372) (15,0.053) (20,0.07395) (25,0.10214) (30,0.11822) (35,0.13923) (40,0.16941) (45,0.19807)}; 
            \addplot+[bar shift=3] coordinates {(0,0.0186) (5,0.02185) (10,0.0266) (15,0.02948) (20,0.03522) (25,0.04307) (30,0.05313) (35,0.06461) (40,0.08045) (45,0.0962)}; 
            \addplot+[bar shift=3.5] coordinates {(0,0.04947) (5,0.05014) (10,0.07241) (15,0.11067) (20,0.15184) (25,0.18552) (30,0.20785) (35,0.2421) (40,0.28657) (45,0.35751)}; 
            \addplot+[bar shift=4] coordinates {(0,0.02272) (5,0.03672) (10,0.06458) (15,0.06902) (20,0.08369) (25,0.10387) (30,0.12726) (35,0.15879) (40,0.2161) (45,0.26859)}; 
            \addplot+[bar shift=4.5] coordinates {(0,0.04979) (5,0.06068) (10,0.0813) (15,0.11383) (20,0.14005) (25,0.16611) (30,0.2071) (35,0.2452) (40,0.30371) (45,0.38494)}; 
            \legend{OLS, TLS|c=1, GMC|c=1, TLS|c=0.6, GMC|c=0.6, TLS|c=0.4, GMC|c=0.4, TLS|c=0.2, GMC|c=0.2}

            \nextgroupplot[
                bar width=4.5pt,
                width=0.12\linewidth,
                height=0.15\textheight,
                symbolic x coords={Average,},
                xlabel={Average},
                xmajorticks=false,
                ytick={0, 0.02, ..., 0.1001},
                yticklabels={0, 2, ..., 10},
            ]
            \addplot+ coordinates {(Average,0.1052)}; 
            \addplot+ coordinates {(Average,0.0437)}; 
            \addplot+ coordinates {(Average,0.0699)}; 
            \addplot+ coordinates {(Average,0.0287)}; 
            \addplot+ coordinates {(Average,0.115)}; 
            \addplot+ coordinates {(Average,0.0266)}; 
            \addplot+ coordinates {(Average,0.1009)}; 
            \addplot+ coordinates {(Average,0.0371)}; 
            \addplot+ coordinates {(Average,0.083)}; 

            \nextgroupplot[
                width=0.74\linewidth,
                height=0.15\textheight,
                bar width=0.4,
                xmax=51,
                xmajorgrids,
                xmin=-1,
                xlabel={Distance / \si{\metre}},
                xtick={0, 5, ..., 50.0001},
                ymax=0.48,
                ytick={0, 0.05, ..., 0.4501},
                yticklabels={0, , 10, , 20, , 30, , 40, },
            ]
            \addplot+[bar shift=0.5] coordinates {(0,0.06538) (5,0.08472) (10,0.11027) (15,0.13874) (20,0.18597) (25,0.23389) (30,0.26491) (35,0.295) (40,0.33482) (45,0.36985)}; 
            \addplot+[bar shift=1] coordinates {(0,0.03813) (5,0.03817) (10,0.04689) (15,0.0529) (20,0.05984) (25,0.06597) (30,0.07054) (35,0.07627) (40,0.08648) (45,0.0977)}; 
            \addplot+[bar shift=1.5] coordinates {(0,0.04967) (5,0.05651) (10,0.07179) (15,0.08901) (20,0.11637) (25,0.14502) (30,0.1691) (35,0.19428) (40,0.23383) (45,0.27285)}; 
            \addplot+[bar shift=2] coordinates {(0,0.0237) (5,0.02501) (10,0.03107) (15,0.03438) (20,0.03883) (25,0.04385) (30,0.04848) (35,0.05512) (40,0.06645) (45,0.07839)}; 
            \addplot+[bar shift=2.5] coordinates {(0,0.07823) (5,0.08531) (10,0.11709) (15,0.15865) (20,0.21272) (25,0.27836) (30,0.34386) (35,0.38735) (40,0.43671) (45,0.46988)}; 
            \addplot+[bar shift=3] coordinates {(0,0.0204) (5,0.02346) (10,0.02812) (15,0.03036) (20,0.03521) (25,0.04386) (30,0.05083) (35,0.05922) (40,0.07259) (45,0.08612)}; 
            \addplot+[bar shift=3.5] coordinates {(0,0.06769) (5,0.07739) (10,0.1026) (15,0.13596) (20,0.18148) (25,0.23187) (30,0.2872) (35,0.32307) (40,0.36399) (45,0.40948)}; 
            \addplot+[bar shift=4] coordinates {(0,0.02117) (5,0.02873) (10,0.04513) (15,0.04665) (20,0.05452) (25,0.07391) (30,0.09276) (35,0.11482) (40,0.15528) (45,0.19977)}; 
            \addplot+[bar shift=4.5] coordinates {(0,0.05625) (5,0.06912) (10,0.08702) (15,0.10575) (20,0.13723) (25,0.1663) (30,0.18647) (35,0.20529) (40,0.24135) (45,0.28255)}; 
        \end{groupplot}
    \end{tikzpicture}
    \caption{
        Abs.~height error between estimated ground surface and ground validation points based on optimization using only labeled ground points (top) and including all points (bottom).
        Left: Average of all validation points.
        Right: Averaged within \SI{5}{\metre} intervals of distance to the sensor.
    }\label{fig:robustifier_influence}
        \begin{tikzpicture}
            \begin{groupplot}[
                    enlarge x limits=0,
                    group style={
                            group size=2 by 1,
                            horizontal sep=0.9cm,
                            ylabels at=edge left,
                        },
                    legend style={
                            anchor=north west,
                            at={(0.03,0.93)},
                            legend columns=-1,
                            nodes={scale=0.75, transform shape},
                        },
                    ybar=1pt,
                    ylabel={Abs.~error / \si{\cm}},
                    ymin=0,
                ]
                \nextgroupplot[
                    bar width=6.5pt,
                    symbolic x coords={Average},
                    width=0.12\linewidth,
                    height=0.15\textheight,
                    xlabel=Average,
                    xmajorticks=false,
                    ytick={0, 0.01, ..., 0.0301},
                    yticklabels={0, 1, ..., 3},
                ]
                \addplot+ coordinates {(Average,0.0281)}; 
                \addplot+ coordinates {(Average,0.0266)}; 
                \addplot+ coordinates {(Average,0.0276)}; 
                \addplot+ coordinates {(Average,0.0374)}; 
                \nextgroupplot[
                    bar width=7pt,
                    width=0.74\linewidth,
                    height=0.15\textheight,
                    xlabel={Distance / \si{\metre}},
                    xmajorgrids,
                    xmax=51,
                    xmin=-1,
                    xtick={0, 5, ..., 50.0001},
                    ytick={0, 0.02, ..., 0.1501},
                    yticklabels={0, 2, ..., 12},
                ]
                \addplot+[bar shift=1] coordinates {(0,0.01994) (5,0.02444) (10,0.03076) (15,0.03298) (20,0.03869) (25,0.04744) (30,0.05609) (35,0.06557) (40,0.07896) (45,0.09177)}; 
                \addplot+[bar shift=2] coordinates {(0,0.0204) (5,0.02346) (10,0.02812) (15,0.03036) (20,0.03521) (25,0.04386) (30,0.05083) (35,0.05922) (40,0.07259) (45,0.08612)}; 
                \addplot+[bar shift=3] coordinates {(0,0.02216) (5,0.02367) (10,0.02937) (15,0.03261) (20,0.03713) (25,0.04548) (30,0.05303) (35,0.062) (40,0.07591) (45,0.09125)}; 
                \addplot+[bar shift=4] coordinates {(0,0.03136) (5,0.03074) (10,0.03974) (15,0.04504) (20,0.05331) (25,0.06706) (30,0.07993) (35,0.09412) (40,0.11505) (45,0.1339)}; 
                \legend{\(r_{\text{asymm}}=2.5\), \(r_{\text{asymm}}=2\), \(r_{\text{asymm}}=1.5\), \(r_{\text{asymm}}=1\)}
            \end{groupplot}
        \end{tikzpicture}
        \caption{
            Abs.~ground point error of \ac{TLS} estimator for different asymmetry ratios.
            Left: Average of all validation points.
            Right: Averaged within \SI{5}{\metre} intervals of distance to the sensor.
        }\label{fig:abs_error_asymmetry}
    \begin{tikzpicture}
        \begin{groupplot}[
                enlarge x limits=0,
                group style={
                        group size=2 by 1,
                        horizontal sep=0.9cm,
                        ylabels at=edge left,
                    },
                legend style={
                        anchor=north west,
                        at={(-0.25,-0.3)},
                        legend columns=5,
                        nodes={scale=1.0, transform shape},
                    },
                ybar=1pt,
                ylabel={Abs.~error / \si{\cm}},
                ymin=0,
            ]
            \nextgroupplot[
                bar width=2.5pt,
                symbolic x coords={Average},
                width=0.12\linewidth,
                height=0.15\textheight,
                xlabel=Average,
                xmajorticks=false,
                ytick={0, 0.01, ..., 0.0601},
                yticklabels={0, 1, ..., 6},
            ]
            \addplot+ coordinates {(Average,0.0225)}; 
            \addplot+ coordinates {(Average,0.0245)}; 
            \addplot+ coordinates {(Average,0.0266)}; 
            \addplot+ coordinates {(Average,0.0372)}; 
            \addplot+ coordinates {(Average,0.0378)}; 
            \addplot+ coordinates {(Average,0.0385)}; 
            \addplot+ coordinates {(Average,0.0532)}; 
            \addplot+ coordinates {(Average,0.0534)}; 
            \addplot+ coordinates {(Average,0.054)}; 
            \legend{\(d_{\text{C}}=\SI{2}{\metre}\) | \(w_{\text{S}}=1\), \(d_{\text{C}}=\SI{2}{\metre}\) | \(w_{\text{S}}=2\), \(d_{\text{C}}=\SI{2}{\metre}\) | \(w_{\text{S}}=10\), \(d_{\text{C}}=\SI{5}{\metre}\) | \(w_{\text{S}}=2\), \(d_{\text{C}}=\SI{5}{\metre}\) | \(w_{\text{S}}=5\), \(d_{\text{C}}=\SI{5}{\metre}\) | \(w_{\text{S}}=10\), \(d_{\text{C}}=\SI{10}{\metre}\) | \(w_{\text{S}}=1\), \(d_{\text{C}}=\SI{10}{\metre}\) | \(w_{\text{S}}=5\), \(d_{\text{C}}=\SI{10}{\metre}\) | \(w_{\text{S}}=10\)}
            \nextgroupplot[
                bar width=0.4,
                width=0.74\linewidth,
                height=0.15\textheight,
                xlabel={Distance / \si{\metre}},
                xmajorgrids,
                xmax=51,
                xmin=-1,
                xtick={0, 5, ..., 50.0001},
                ytick={0, 0.02, ..., 0.1201},
                yticklabels={0, 2, ..., 12},
            ]
            \addplot+[bar shift=0.5] coordinates {(0,0.01927) (5,0.02085) (10,0.02346) (15,0.02401) (20,0.0258) (25,0.03071) (30,0.03465) (35,0.04045) (40,0.05075) (45,0.06664)}; 
            \addplot+[bar shift=1] coordinates {(0,0.01977) (5,0.02188) (10,0.02556) (15,0.02736) (20,0.03088) (25,0.03784) (30,0.04425) (35,0.0521) (40,0.06513) (45,0.07813)}; 
            \addplot+[bar shift=1.5] coordinates {(0,0.0204) (5,0.02346) (10,0.02812) (15,0.03036) (20,0.03521) (25,0.04386) (30,0.05083) (35,0.05922) (40,0.07259) (45,0.08612)}; 
            \addplot+[bar shift=2] coordinates {(0,0.03139) (5,0.03465) (10,0.04256) (15,0.0405) (20,0.04152) (25,0.04469) (30,0.04731) (35,0.05191) (40,0.06137) (45,0.07494)}; 
            \addplot+[bar shift=2.5] coordinates {(0,0.03136) (5,0.03516) (10,0.04282) (15,0.04079) (20,0.04274) (25,0.0473) (30,0.05203) (35,0.05876) (40,0.07083) (45,0.08408)}; 
            \addplot+[bar shift=3] coordinates {(0,0.03135) (5,0.03549) (10,0.0433) (15,0.04164) (20,0.04458) (25,0.05061) (30,0.05602) (35,0.06353) (40,0.0764) (45,0.08937)}; 
            \addplot+[bar shift=3.5] coordinates {(0,0.03793) (5,0.04716) (10,0.06332) (15,0.06492) (20,0.07082) (25,0.07875) (30,0.08692) (35,0.09061) (40,0.09679) (45,0.11012)}; 
            \addplot+[bar shift=4] coordinates {(0,0.03813) (5,0.04748) (10,0.06428) (15,0.06553) (20,0.07124) (25,0.07707) (30,0.08162) (35,0.08524) (40,0.09172) (45,0.10255)}; 
            \addplot+[bar shift=4.5] coordinates {(0,0.03823) (5,0.04783) (10,0.06536) (15,0.06673) (20,0.07235) (25,0.07828) (30,0.08198) (35,0.08594) (40,0.09315) (45,0.10469)}; 
        \end{groupplot}
    \end{tikzpicture}
    \caption{
        Abs.~ground point error of \ac{TLS} estimator for different control point distances \(d_{\text{C}}\) and smoothness weights \(w_{\text{S}}\).
        Left: Average of all validation points.
        Right: Averaged within \SI{5}{\metre} intervals of distance to the sensor.
    }\label{fig:abs_error_control_points}
\end{figure*}
We observe that the \ac{TLS} method yields the best results in presence of outliers.
An optimal error threshold for the \ac{TLS} method seems to be in the range of \SIrange{20}{60}{\cm}.
The \ac{GMC} approach does not always yield better results than the \ac{OLS} baseline.
This may be because the \ac{GMC} method does not converge within \num{5} iterations.

\subsection{Asymmetric Cost}\label{subsec:asymmetry}
\Cref{fig:abs_error_asymmetry} summarizes the absolute errors between ground validation points and the estimated ground surface at different asymmetry ratios.

We observe that the \ac{TLS} estimator yields the lowest ground surface errors for \(1.5 < r_{\text{asymm}} < 2.5\).

\subsection{Control Point Distance vs.~Smoothness Weight}\label{subsec:control_points}
\Cref{fig:abs_error_control_points} compares the abs.~errors between estimated ground surface and ground validation points for different control point distances and smoothness weights.

In general, we observe that smaller control point distances and smoothness weights yield smaller errors.
The influence of the smoothness weight increases with decreasing measurement density, often in areas far away from the sensor.

%% file: content/validation.tex
\section{Validation on Experimental Vehicle}\label{sec:validation}

We implemented the ground surface estimation on our experimental vehicle equippped with four Velodyne VLP16 \acsp{LIDAR} on the roof corners and one Velodyne VLS128 \cite{AlphaPrime} on the roof center.

\subsection{Experimental Vehicle}\label{sec:experimental_vehicle}

\begin{figure}
    \centering
    \def\svgwidth{\columnwidth}
        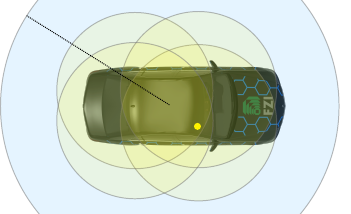
        \caption{
            Sensor setup top-view on the experimental vehicle.
            Not to scale.\\
            Yellow: VLP16, blue: VLS128, red: LUX4L.}\label{fig:bertha_sensors}
\end{figure}

Our experimental vehicle is an automated Mercedes-Benz E class limousine.
As depicted in \cref{fig:bertha_sensors} it has mounted four Velodyne VLP16 \acsp{LIDAR} on the roof corners, one Velodyne VLS128 \acs{LIDAR} on the roof center and an Ibeo LUX4L \acs{LIDAR} below the front sign plate.

Computations are made on a general-purpose PC with an AMD EPYC 7702P 64-Core processor and 256GB RAM.
The PC has two \acsp{GPU}, an NVidia TITAN X and a Titan V.

\subsection{Validation}
The optimization time mainly depends on the number of iterations, point measurements and control points.
The range sensor setup generates approximately \num{360000} point measurements per scan at a scan rate of \SI{10}{\Hz}, which results in an approximate point measurement rate of \SI{3.6}{\mega\Hz}.
The optimization time is proportional to the number of iterations.
However, as we receive sequential measurements, we initialize the ground surface with the last optimization result and perform only one iteration.
Thus, we adapt the number of control points so that the optimization satisfies soft real-time constraints.

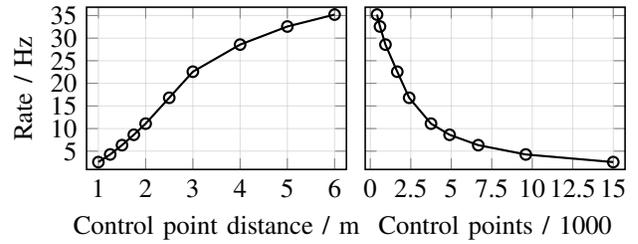
\begin{figure}
    \begin{tikzpicture}
        \begin{groupplot}[
                grid=major,
                group style={
                        group size=2 by 1,
                        horizontal sep=0.25cm,
                        ylabels at=edge left,
                    },
                ylabel={Rate / \si{\Hz}},
            ]
            \nextgroupplot[
                width=0.4\linewidth,
                xlabel={Control point distance / \si{\metre}},
                xtick={0, 1, ..., 6},
                ytick={0, 5, ..., 35},
            ]
            \addplot[black, fill=none, mark=o] coordinates {(1.0,2.582) (1.25,4.27) (1.5,6.329) (1.75,8.621) (2.0,11.086) (2.5,16.807) (3.0,22.573) (4.0,28.571) (5.0,32.573) (6.0,35.211)};
            \nextgroupplot[
                width=0.4\linewidth,
                xlabel={Control points / \num{1000}},
                xtick={15000, 12500, ..., 0},
                xticklabels={15, 12.5, ..., 0},
                ytick={0, 5, ..., 35},
                yticklabels={,,},
            ]
            \addplot[black, fill=none, mark=o] coordinates {(15000.0,2.582) (9600.0,4.27) (6666.667,6.329) (4897.959,8.621) (3750.0,11.086) (2400.0,16.807) (1666.667,22.573) (937.5,28.571) (600.0,32.573) (416.667,35.211)};
        \end{groupplot}
    \end{tikzpicture}
    \caption{
        Ground surface estimation processing rate on the experimental vehicle depending on the control points within a \SI{150}{\metre}\texttimes\SI{100}{\metre} area.
    }\label{fig:computation_times_bertha}
\end{figure}

\Cref{fig:computation_times_bertha} summarizes the computation rates on the experimental vehicle depending on the control points within a \SI{150}{\metre}\texttimes\SI{100}{\metre} area.
For example, if the ground surface estimation should process measurements at a rate of at least \SI{10}{\Hz}, the number of control points should be less than \num{3750} or in other words, the control point distance in this area should be at least \SI{2}{\metre}.

\begin{figure*}[t!]
    \centering
    \includegraphics[width=\linewidth]{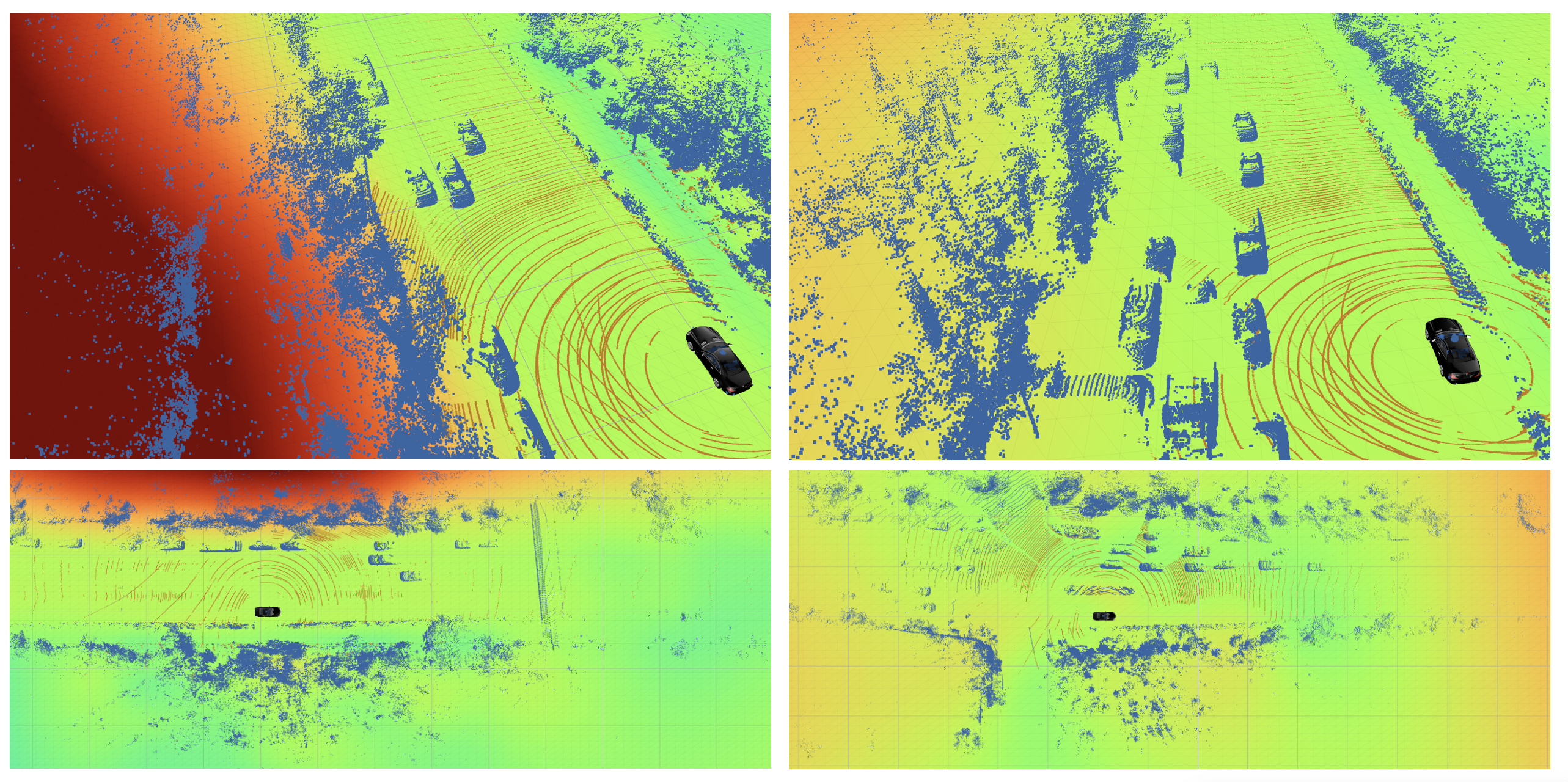}
    \caption{
        Point set from full \SI{360}{\degree} scans of all \acsp{LIDAR} on the experimental vehicle and estimated ground surface.
        The point set is colorized by the distance of points to the ground surface with brown denoting a distance of less than \SI{10}{\cm} and blue above \SI{10}{\cm}.
        The ground surface is colorized by its height relative to the vehicle reference frame.
    }\label{fig:bertha_qualitative}
\end{figure*}

\Cref{fig:bertha_qualitative} shows the point set from full \SI{360}{\degree} scans of all \acsp{LIDAR} mounted on the experimental vehicle together with the estimated ground surface on a drive through Karlsruhe, Germany.
We observe that the ground surface can be accurately estimated.
Based on the resulting ground surface, we are able to distinguish between ground / non-ground points by applying a simple distance-based classifier.

%% file: fig/diss/bertha_setup.pdf_tex
\begingroup%
  \makeatletter%
  \providecommand\color[2][]{%
    \errmessage{(Inkscape) Color is used for the text in Inkscape, but the package 'color.sty' is not loaded}%
    \renewcommand\color[2][]{}%
  }%
  \providecommand\transparent[1]{%
    \errmessage{(Inkscape) Transparency is used (non-zero) for the text in Inkscape, but the package 'transparent.sty' is not loaded}%
    \renewcommand\transparent[1]{}%
  }%
  \providecommand\rotatebox[2]{#2}%
  \newcommand*\fsize{\dimexpr\f@size pt\relax}%
  \newcommand*\lineheight[1]{\fontsize{\fsize}{#1\fsize}\selectfont}%
  \ifx\svgwidth\undefined%
    \setlength{\unitlength}{100.96674326bp}%
    \ifx\svgscale\undefined%
      \relax%
    \else%
      \setlength{\unitlength}{\unitlength * \real{\svgscale}}%
    \fi%
  \else%
    \setlength{\unitlength}{\svgwidth}%
  \fi%
  \global\let\svgwidth\undefined%
  \global\let\svgscale\undefined%
  \makeatother%
  \begin{picture}(1,0.60819271)%
    \lineheight{1}%
    \setlength\tabcolsep{0pt}%
    \put(0,0){\includegraphics[width=\unitlength,page=1]{bertha_setup.pdf}}%
    \put(0.06550795,0.45894188){\color[rgb]{0,0,0}\makebox(0,0)[lt]{\lineheight{1.25}\smash{\begin{tabular}[t]{l}\SI{200}{\metre}\end{tabular}}}}%
    \put(0,0){\includegraphics[width=\unitlength,page=2]{bertha_setup.pdf}}%
    \put(0.87028761,0.25267663){\color[rgb]{0,0,0}\makebox(0,0)[lt]{\lineheight{1.25}\smash{\begin{tabular}[t]{l}\SI{50}{\metre}\end{tabular}}}}%
    \put(0,0){\includegraphics[width=\unitlength,page=3]{bertha_setup.pdf}}%
    \put(0.2705578,0.10509977){\color[rgb]{0,0,0}\makebox(0,0)[lt]{\lineheight{1.25}\smash{\begin{tabular}[t]{l}\SI{100}{\metre}\end{tabular}}}}%
    \put(0,0){\includegraphics[width=\unitlength,page=4]{bertha_setup.pdf}}%
  \end{picture}%
\endgroup%

%% file: content/conclusion.tex
\section{Conclusion}\label{sec:conclusion}

We proposed a method for ground surface estimation from noisy range measurements represented as point sets.
In our approach, we modeled the ground surface as \ac{UBS}.
\Acp{UBS} implicitly impose smoothness and are insensitive to locally varying measurement densities.
With robust optimization techniques and the \ac{UBS} surface model, we were able to accurately estimate the ground surface in a wide distance range.
Using this ground surface estimate, we are able to distinguish between ground and obstacle surface reflections so that we can model the traffic scene relative to the ground surface.